\documentclass[letterpaper, 10 pt, conference]{ieeeconf}  

\IEEEoverridecommandlockouts                              
\usepackage{graphicx}
\usepackage{multicol}
\usepackage{subcaption}
\usepackage{listings}
\usepackage{algorithm}
\usepackage{algorithmicx}
\usepackage{algpseudocode}
\usepackage{pifont}

\usepackage{amsmath} 
\usepackage{fontenc}
\usepackage{inputenc}
\usepackage{amssymb} 
\usepackage{cite}
\usepackage{placeins}
\graphicspath{{images/}}
\usepackage{hyperref}
\hypersetup{
    colorlinks = true,
    linkcolor = blue,
}
\usepackage[nolist,nohyperlinks]{acronym}
\title{\LARGE \bf
iCub: Learning Emotion Expressions using Human Reward
}

\author{Nikhil Churamani, Francisco Cruz, Sascha Griffiths and Pablo Barros 
\thanks{The authors are with Knowledge Technology, Department of Informatics, University of Hamburg, Germany.{\tt\scriptsize Email: 
        \{5churama, cruz, griffiths, barros\}@informatik.uni-hamburg.de}}
\thanks{Published in the \href{https://www2.informatik.uni-hamburg.de/wtm/SocialRobotsWorkshop2016/index.php}{Workshop on Bio-inspired Social Robot Learning in Home Scenarios},  IEEE/RSJ International Conference on Intelligent Robots and Systems (IROS), Daejeon, Korea (2016)}
}

\begin{document}

\maketitle
\thispagestyle{empty}
\pagestyle{empty}

%%%%%%%%%%%%%%%%%%%%%%%%%%%%%%%%%%%%%%%%%%%%%%%%%%%%%%%%%%%%%%%%%%%%%%%%%%%%%%%%
\begin{abstract}

The purpose of the present study is to learn emotion expression representations for artificial agents using reward shaping mechanisms. The approach takes inspiration from the TAMER framework for training a \ac{MLP} to learn to express different emotions on the iCub robot in a human-robot interaction scenario. The robot uses a combination of a \ac{CNN} and a \ac{SOM} to recognise an emotion and then learns to express the same using the \ac{MLP}. The objective is to teach a robot to respond adequately to the user's perception of emotions and learn how to express different emotions.

\end{abstract}

%%%%%%%%%%%%%%%%%%%%%%%%%%%%%%%%%%%%%%%%%%%%%%%%%%%%%%%%%%%%%%%%%%%%%%%%%%%%%%%%
\section{Introduction}
\label{sec:introduction}

With the advances in human-robot interaction research, artificial agents are slowly but surely becoming an integral part of human life. We are going to interact with these agents on a daily basis, in one form or the other. As our interaction and thus inadvertently, our dependency on agents increases, these agents need to blend into the social environments surrounding them, as naturally as possible. Thus, agents need to be able to factor in emotions and sentiments while dealing with their human-centred environment so as to make well-informed decisions. 

This study makes use of the iCub robot head and its on-board capabilities to encode different emotional expressions for seven ``Universal Emotions'', as proposed by Ekman \cite{ekman1992argument} viz. \textit{Anger}, \textit{Disgust}, \textit{Fear}, \textit{Happiness}, \textit{Sadness} and \textit{Surprise} along with \textit{Neutral} as an additional emotional state to accommodate for the absence of any emotion. 

Also, this work explores \ac{RL} strategies, particularly in the context of \textit{reward shaping} \cite{sutton1998reinforcement} approaches to train an \ac{MLP} to learn different emotion expression states for perceived emotions. The agent makes use of the on-board camera to capture an image of the face of the user and uses the \ac{CNN}, to come up with a feature representation for the face image. These feature representations are then used to train the \ac{SOM}, as presented in \cite{barros2016}, forming clusters representing each of the aforementioned seven emotions. The \ac{MLP} is then used to come up with the most appropriate expression representation related to an action to be used for the emotion. This work builds on the work by Barros et. al \cite{barros2016} and uses the underlying \ac{CNN} and \ac{SOM} model to enhance the proposed system.

The implementation is inspired by the TAMER \cite{knox2008tamer,knox2012learning,knox2013training} algorithm for modelling a ``\textit{learning by human reward}'' approach so as to train the \ac{MLP}. 

%%%%%%%%%%%%%%%%%%%%%%%%%%%%%%%%%%%%%%%%%%%%%%%%%%%%%%%%%%%%%%%%%%%%%%%%%%%%%%%%

\section{Emotion Expression using icub}
\label{sec:icub}
\begin{figure}
\centering
\includegraphics[width=0.3\textwidth]{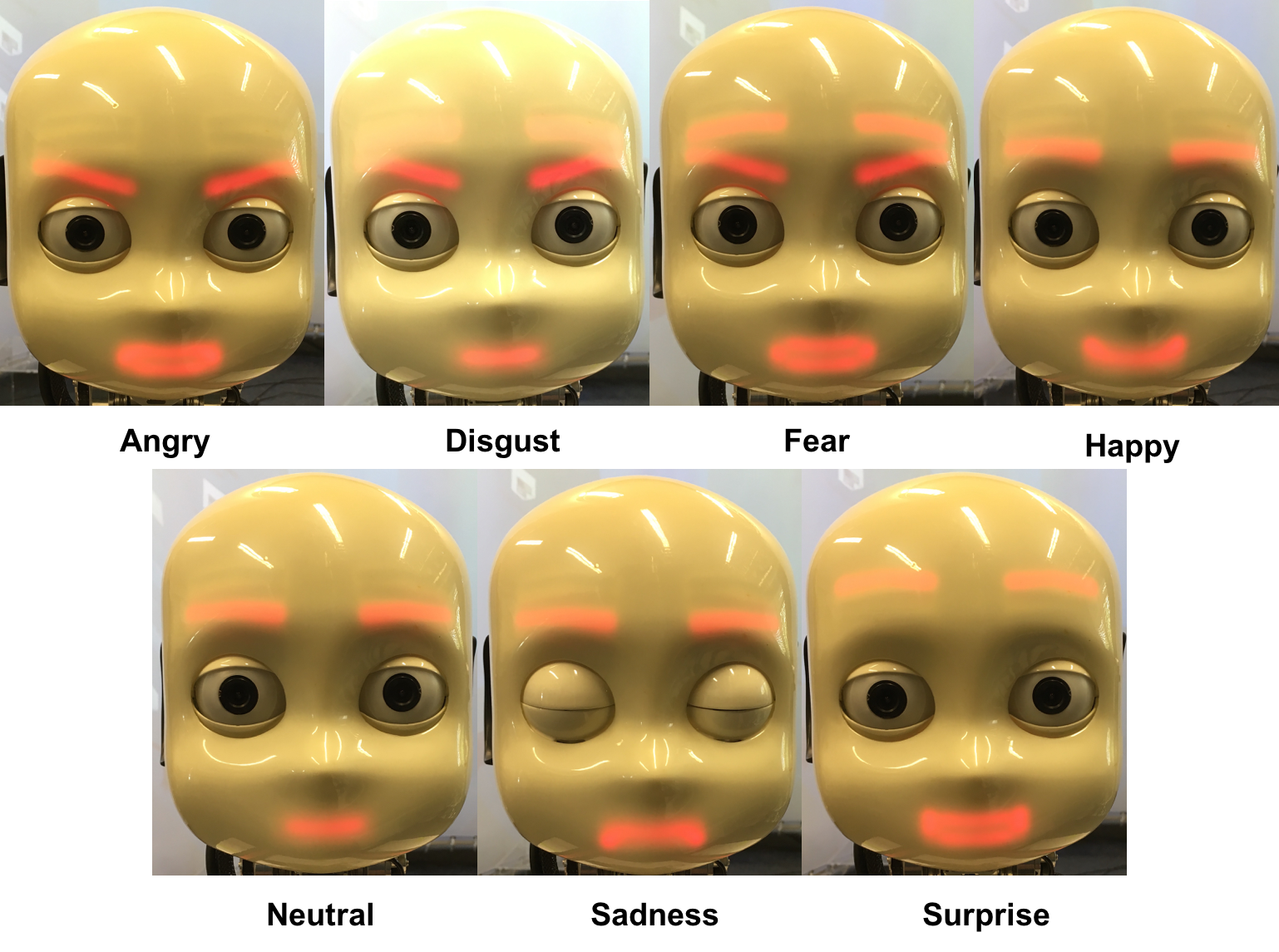}
\caption{iCub Robot Head encoding emotions}
\label{fig:icubface}
% \vspace*{-3mm}
\end{figure}

In the context of this study, we make use of the iCub robot head \cite{icub2015} which uses multiple underlying electronic control boards corresponding to LEDs representing the left and the right eyebrows and the mouth. The iCub robot head uses an \textit{emotionInterface} module which is able to map different emotion states to expression representations via the LEDs available on the face. Different subsystems like the mouth, eye-lids, left eyebrow, right eyebrow or the whole system can be used alternatively to encode different emotion states by setting LED flags for these subsystems. The encodings for emotion expressions used in this study are shown in Fig.\ref{fig:icubface}.

The study assumes that there is a unique response, taking the form of an ``expression selection'', to each input emotion state and thus, one particular facial representation shall uniquely identify a particular emotion. Forming a particular facial representation (using a combination of lights, selecting them one by one) can also be broken down into subtasks which could be learned as well. In this study, a particular combination of lights is selected as the lowest abstraction level for action selection for the sake of simplicity.

%%%%%%%%%%%%%%%%%%%%%%%%%%%%%%%%%%%%%%%%%%%%%%%%%%%%%%%%%%%%%%%%%%%%%%%%%%%%%%%%
\section{Proposed Model}
\label{sec:impl}

\begin{figure*}[ht]
\centering
\includegraphics[width=0.9\textwidth]{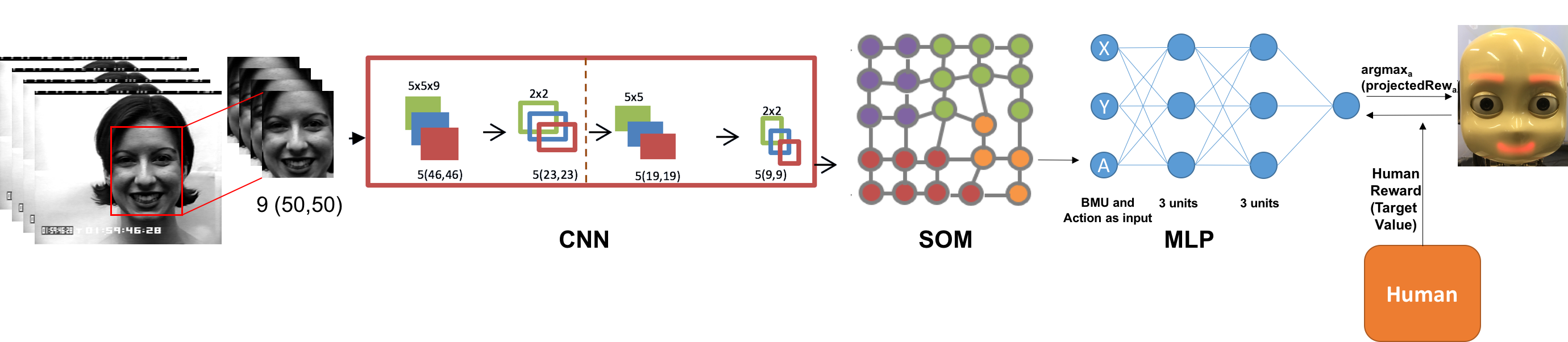}
\caption{Complete Model with \ac{CNN} face representation, \ac{SOM} clusters and the \ac{MLP} predicting a reward and selecting an action learning from human reward}
\label{fig:complete}
% \vspace*{-3mm}

\end{figure*}

The motivation behind this study is to explore the possibility of training artificial agents to perform actions, while interacting with a human user. In the context of emotions, different people express and perceive emotions differently and thus, the agent needs to adapt to this variance and customise its learning depending on the user. 

The iCub captures an image of the user's expression which is fed to a pre-trained \ac{CNN} giving it a feature vector representation. The face feature representations are then fed to the \ac{SOM} where it is expected that clusters emerge, each pertaining to a particular emotion. User interactions are modelled by taking the \ac{BMU}  from the \ac{SOM} for training an \ac{MLP} to predict the action which best mimics the user's expression. The user rewards the robot's action and the \ac{MLP} is trained to select the correct expression by learning to predict this reward. The \ac{CNN} is trained using the Cohn-Kanade dataset \cite{lucey2010extended}, in this case consisting of approximately 1000 different images. The \ac{SOM} on the other hand is trained each time for a specific user. 
\begin{figure}
\centering
\includegraphics[width=0.408\textwidth]{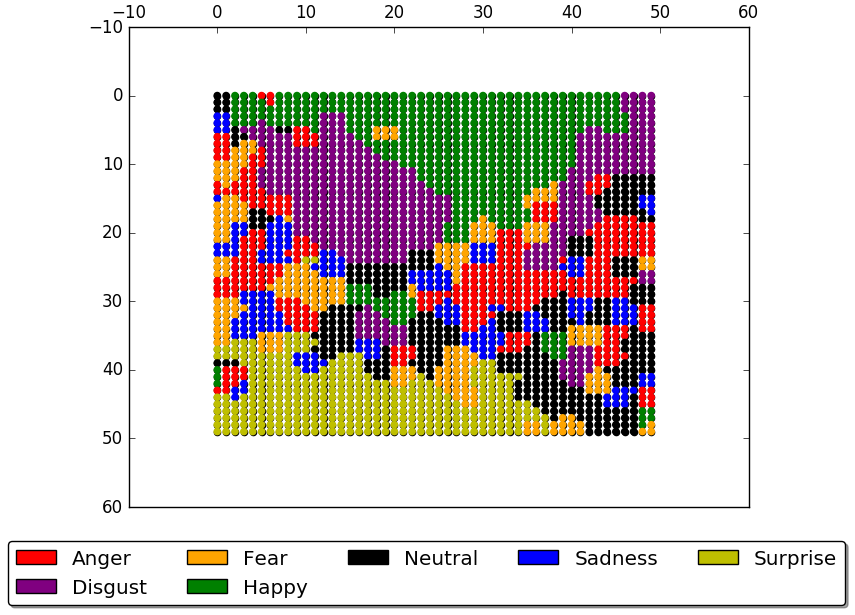}
\caption{\ac{SOM} Cluster Map}
\label{fig:somsample}
% \vspace*{-4mm}
\end{figure}

The \ac{CNN} consists of two convolution layers (with max-pooling) also making use of L1 and L2 normalisation for each layer respectively. The resulting feature vector representation is used to train the \ac{SOM}, forming clusters for different emotions (see Fig. \ref{fig:somsample}). For each user interaction, the \ac{BMU} is computed (position) and fed to the \ac{MLP} along with a possible action (expression representation) resulting in a predicted reward value ($ r$ $ \epsilon$ $[-2,2]$). The agent then tries all possible actions and chooses the action with the highest reward value and performs that action. Fig. \ref{fig:complete} depicts the complete model.

Once the iCub performs an action, the user is expected to reward it, thus giving it a target value to achieve. This is done by asking the user to mimic the robot giving it information about how much the action performed differs from the intended action. The iCub again captures the user mimicking the agent and evaluates the \ac{BMU} for the reward. The reward value ($ r$ $ \epsilon$ $[-2,2]$) is computed by using the \textit{Normalised Euclidean Distance} between the \ac{BMU}s and thresholding it i.e. the lesser the distance between the \ac{BMU}s, the more positive the reward. The aim of the \ac{MLP} learning algorithm is thus, to approximate an optimal reward function which is able to estimate the user's reward for each emotion-expression pair. 
% * <5churama@informatik.uni-hamburg.de> 2016-09-12T18:19:48.752Z:
%
% The reward value ($ r$ $ \epsilon$ $[-2,2]$) is computed by using the \textit{Normalised Euclidean Distance} between the \ac{BMU}s and thresholding it i.e. lesser the distance between the \ac{BMU}s, more positive the reward. 
%
% ^.
%%%%%%%%%%%%%%%%%%%%%%%%%%%%%%%%%%%%%%%%%%%%%%%%%%%%%%%%%%%%%%%%%%%%%%%%%%%%%%%%

\section{Preliminary Results}

\begin{figure}
\centering
  \begin{subfigure}[t]{0.24\textwidth}	
    \includegraphics[width=1\textwidth]{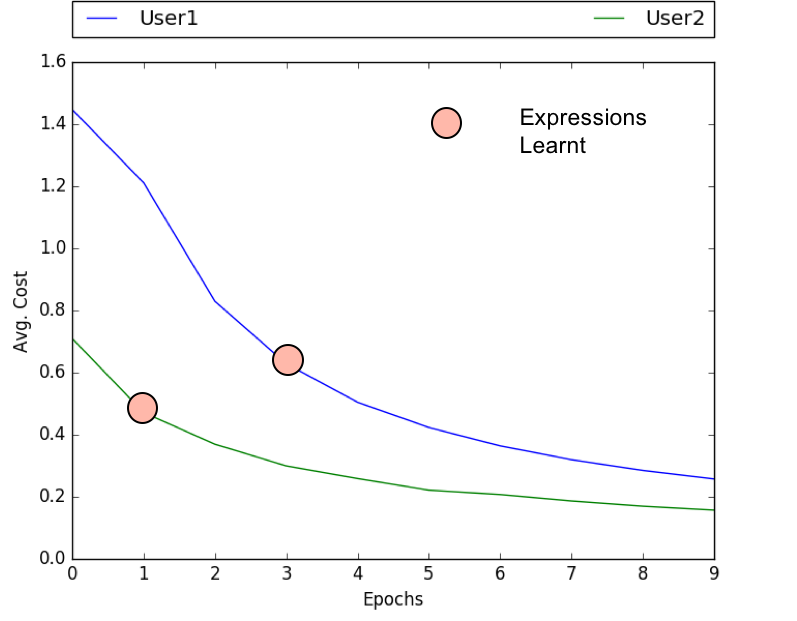}
    \caption{Users with no prior knowledge about the system}
    \label{fig:exp1}
  \end{subfigure}
  \begin{subfigure}[t]{0.22\textwidth}
    \includegraphics[width=1\textwidth]{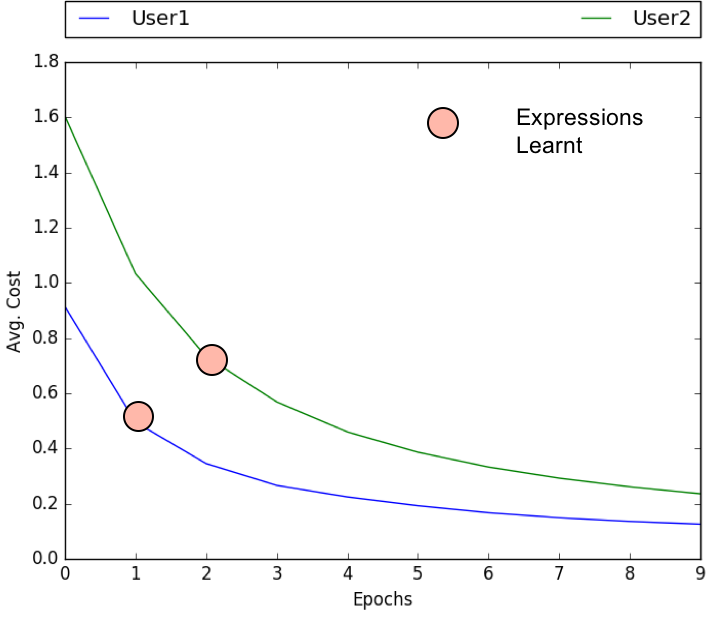}
    \caption{Users with prior knowledge about the system }
    \label{fig:exp2}
  \end{subfigure}
  \caption{Preliminary Results: Each epoch corresponding to 100 interactions for calculating the Avg. cost}
  \label{fig:result}
%   \vspace*{-4mm}
\end{figure}

Some preliminary studies were performed with users for the model providing positive and motivating results (see Fig. \ref{fig:result}). Although the results were promising and significantly reduced the time needed for training, the method still required more than 100 interactions per user to learn meaningful expressions. This number is expected to decrease with improvements in training methodologies and by collecting more data for training. 

% \FloatBarrier

%%%%%%%%%%%%%%%%%%%%%%%%%%%%%%%%%%%%%%%%%%%%%%%%%%%%%%%%%%%%%%%%%%%%%%%%%%%%%%%%
\begin{acronym}
\acro{MLP}{Multilayer Perceptron}
\acro{CNN}{Convolutional Neural Network}
\acro{SOM}{Self-organising Map}
\acro{BMU}{Best Matching Unit}
\acro{TAMER}{Training an Agent Manually via Evaluative Reinforcement}
\acro{RL}{Reinforcement Learning}
\end{acronym}
\bibliographystyle{ieeetr}
\bibliography{study}

\end{document}